\documentclass[10pt,twocolumn,letterpaper]{IEEEtran}
\makeatletter
\def\ps@headings{%
\def\@oddhead{\mbox{}\scriptsize\rightmark \hfil \thepage}%
\def\@evenhead{\scriptsize\thepage \hfil \leftmark\mbox{}}%
\def\@oddfoot{}%
\def\@evenfoot{}}
\makeatother
\IEEEoverridecommandlockouts 
\usepackage{graphicx,latexsym,subfigure,amsmath,epsfig,latexsym,subfigure,amsmath,cite,amssymb}
\usepackage{amsthm,mathrsfs}
\usepackage{booktabs}
\usepackage{graphicx}
\usepackage{epstopdf}
\usepackage{color}
\usepackage{CJK}
\usepackage{cite} 
\usepackage{algorithmic}
\usepackage{mathrsfs}
\usepackage{bm}
\usepackage{float}
\usepackage{setspace}
\usepackage{color}
\usepackage{subfigure}
\usepackage{balance}
\usepackage{dsfont} 
\usepackage{bm}
\usepackage[ruled,linesnumbered]{algorithm2e} 

    \newtheorem{theorem}{\hspace{1em}Theorem}

    \newcommand*{\QEDA}{\hfill\ensuremath{\blacksquare}}  

\SetKwRepeat{DoWhile}{Do}{While}

\begin{document}
\begin{CJK}{GBK}{song}

\title{\huge Sparsity-Aware Intelligent Massive Random Access Control in Open RAN: A Reinforcement Learning Based Approach}
\author{
        Xiao~Tang,~
        Sicong~Liu,~\IEEEmembership{Senior~Member,~IEEE,}
        Xiaojiang~Du,~\IEEEmembership{Fellow,~IEEE,}
        and~Mohsen~Guizani,~\IEEEmembership{Fellow,~IEEE}
\thanks{}
}


\maketitle
\begin{spacing}{1}
\begin{abstract}
  Massive random access of devices in the emerging Open Radio Access Network (O-RAN) brings great challenge to the access control and management.
  Exploiting the bursting nature of the access requests, sparse active user detection (SAUD) is an efficient enabler towards efficient access management,
  but the sparsity might be deteriorated in case of uncoordinated massive access requests. To dynamically preserve the sparsity of access requests,
  a reinforcement-learning (RL)-assisted scheme of closed-loop access control utilizing the access class barring technique is proposed,
  where the RL policy is determined through continuous interaction between the RL agent, i.e., a next generation node base (gNB), and the environment.
  The proposed scheme can be implemented by the near-real-time RAN intelligent controller (near-RT RIC) in O-RAN,
  supporting rapid switching between heterogeneous vertical applications, such as mMTC and uRLLC services. Moreover,
  a data-driven scheme of deep-RL-assisted SAUD is proposed to resolve highly complex environments with continuous and high-dimensional state and action spaces,
  where a replay buffer is applied for automatic large-scale data collection. An actor-critic framework is formulated to incorporate the strategy-learning
  modules into the near-RT RIC. Simulation results show that the proposed schemes can achieve superior performance in both access efficiency and user detection accuracy over the
  benchmark scheme for different heterogeneous services with massive access requests.

\end{abstract}

\renewcommand{\thefootnote}{}
\footnote{X. Tang and S. Liu are with the Department of Information and Communication Engineering, School of Informatics, Xiamen University, Xiamen 361005, China (E-mail: liusc@xmu.edu.cn).
X. Du is with the Department of Electrical and Computer Engineering, Stevens Institute of Technology, Hoboken, NJ 07030 USA (E-mail: dxj@ieee.org).
Mohsen Guizani is with Mohamed Bin Zayed University of Artificial Intelligence, United Arab Emirates (E-mail: mguizani@ieee.org). (\emph{Corresponding Author: Sicong Liu)} }

\vspace{-0.05in}
\section{Introduction}\label{sec:intro}
In recent years, international for standardization organizations and global mobile operators have begun to devote themselves to the development and application of the emerging architecture of open radio access network (O-RAN).
Relying on network function virtualization and software-defined network  technologies, O-RAN is aimed at improving the portability and flexibility of RAN,
and creating interoperable and programmable RAN products and services \cite{t1}.
Near-real-time RAN Intelligent Controller (near-RT RIC) is the key to efficient network management in local O-RAN \cite{t2}.

  For heterogeneous services in O-RAN, such as radio fingerprint identification, handover, and traffic steering, various third-party multiple control applications, i.e., xApps,
  are running on the near-RT RIC, providing programmable network management strategies of the next generation node base (gNB) for different scenarios to achieve quality of service (QoS) optimization \cite{t3,t4}.
  It is required by various applications and services, especially mMTC services, that the access of high density and large-scale devices ($10^2/\rm{km}^2$ to $10^7/\rm{km}^2$) to the RAN should be supported, which brings great challenge to random access management \cite{t5,t6}.
  Before the user equipment successfully establishes a connection with the core network, the gNB needs to detect the active users from all possible potential users in order to send a response to the correct user.
  The problem of active user detection (AUD) has been investigated in O-RAN Working Group 3.

    Fortunately, the connection requests of massive users have a bursting and sparse nature, i.e., at a particular time slot, the active users, who need to make a request to access the network,
    occupy only a small proportion of all the potential users residing within this network. This sparse nature makes it possible for an efficient technique, i.e.,
    the compressed sensing (CS)-based AUD, to be feasible, which is also called sparse AUD (SAUD) \cite{t7,t8,t9,t10}. However, the performance of SAUD might be degraded if the sparsity of the access requests is deteriorated \cite{t24}.
    For example, when the number of access requests increase rapidly, or when multiple heterogeneous services are supported simultaneously in the network \cite{tdu}, more conflicts among the active users might occur,
    which breaks the sparsity nature and results in detrimental impact on the accuracy of AUD.

    Hence, it is necessary to design an effective paradigm to properly manage and dynamically control the access requests of the users in the network so that the sparsity of the access requests can be preserved.
    As a standard flow control mechanism in 3GPP, the technique of access class barring (ACB) defines a specific access priority for some certain class of users.
    When the network is overloaded, it can effectively reduce the network burden by setting the ACB factor and imposing priority-based access requests,
    thus improving the performance of SAUD in critical severe conditions \cite{t11,t12}.

    On the other hand, with the emergence of O-RAN, a new yet efficient alternative of intelligent access control is provided for heterogeneous vertical services.
    Utilizing network slicing, an independent O-RAN can be divided into multiple logical networks, so that the near-RT RIC can implement specific access control strategies according to various service requirements.
    For instance, uRLLC services such as internet of vehicles are more sensitive to the reliability of the connection, so they pay more attention to the accuracy of AUD,
    while mMTC services such as internet of things tend to enable more users to access even at the cost of reliability.
    The network should be able to adaptively soft switch between different heterogeneous services, and also deal with the time-varying communication environments in a particular service.

    Therefore, in order to meet the diverse demands, xAPPs deployed on near-RT RIC need to adaptively adjust their configuration, which requires a dynamic closed-loop decision and control scheme.
    Machine learning (ML) and reinforcement learning (RL) technologies can be introduced to O-RAN \cite{t13,t14} to facilitate a data-driven intelligent agent at the gNB with programmable custom logic,
    which can proactively interact with the time-varying environment and dynamically configure the network access control strategy in different services.

    To this end, an RL-based framework is proposed in this paper to be deployed on near-RT RIC to realize dynamic access management through adaptive control of the ACB factor.
    Specifically, an RL-assisted SAUD (RL-SAUD) is devised, which adopts Q-learning to obtain experience from trials and errors and learn appropriate decision-making policies.
    The process of the ACB control and the closed-loop interaction with the environment can be modeled as a Markov decision process (MDP), which can dynamically update the access control strategy of the near-RT RIC.
    In the proposed RL-SAUD scheme, different ACB factors are assigned to different priority-based classes, making a reasonable trade-off between the permitted access probability and the user detection accuracy.

    Furthermore, in order to support effective control towards continuous and high-dimensional state and action spaces,
    a deep reinforcement learning (DRL) assisted SAUD (DRL-SAUD) with Actor-Critic framework is devised to mitigate the impact of the state quantization and the dimensional curse in Q-learning.
    Besides, previous experience is obtained via pre-training and stored in a replay buffer, which is exploited to initialize the parameters of the deep neural networks for more rapid convergence.
    Then, through the closed-loop interaction between the gNB agent and the network environment, the optimal strategy of access control can be achieved in near real time.
    The superiority of the proposed scheme is verified by both theoretical analysis and simulation experiments. Consequently, the main contributions of this work are summarized as follows.

    \begin{itemize}
      \item {\color{black}A massive access control framework based on ACB and SAUD is formulated, which is deployed in the xAPP of the near-RT RIC for O-RAN. The inherent sparsity of access requests is fully exploited to achieve efficient and accurate detection of active users among massive amount of potential users in the network.}
      \item An RL-assisted SAUD (RL-SAUD) is devised to realize dynamic access management through adaptive and closed-loop control of the ACB factor via Q-learning, which can adaptively preserve the sparsity of access requests that may be deteriorated by severe conflicts of excessively massive access requests or heterogeneous services, thus sustaining the accuracy of SAUD.
      \item Furthermore, a DRL-assisted SAUD scheme (DRL-SAUD) with built-in actor-critic module is proposed, where the previous experience is utilized for faster convergence. A data-driven paradigm enabled by training the deep neural networks can resolve the performance limitation due to quantification error and dimensional curse of continuous state and action spaces.
    \end{itemize}

    The rest of this article is organized as follows. Section II reviews the related prior work. The system model of massive random access in the context of O-RAN is presented in Section III.
    The proposed schemes of RL-SAUD and DRL-SAUD for intelligent and dynamic ACB factor and massive access control are described in detail in Section IV and Section V, respectively.
    The theoretical performance analysis is given in Section VI. Simulation results are reported in Section VII, which is followed by the conclusions in Section VIII.

\vspace{0.1in}
\section{Related WORK}\label{sec:rw}
In the contention free random access paradigm, the gNB allocates a specific preamble sequence to the potential users in advance to accelerate the average speed of service recovery \cite{t15}.
Users intended to access the network send their preambles on the physical random access channel, and the gNB detects the active user according to the known assignment relation between the preamble and the user.
This process is called AUD, which aims to detect the correct user to establish a random access connection for information transfer.
O-RAN has been developing based on existing network interface protocols and compatible with more heterogeneous services, where AUD has thus become a key challenge.
Some techniques have been investigated to achieve acceptable AUD performance, such as receive beamforming based blind multi-user detection \cite{t16} and coordinated detection between cells \cite{t17}.

Usually, the number of active users raising access requests is much less than the total number of potential users within the network, which shows a sparse nature in active traffic.
The sparsity of the traffic turns the AUD problem into a sparse recovery problem, which can be represented in a CS framework, i.e., SAUD \cite{t5,t18}.
Thus, some sparse recovery methods such as approximate message passing with block sparse Bayesian learning \cite{t19} and joint sparse support recovery with multiple measurement vectors \cite{t20} can be adopted,
reducing the complexity of AUD in case of massive access requests. However, due to the surge of data traffic in O-RAN, the performance of SAUD might still suffer from degradation.
Compared with pure hardware processing, a better approach is to preserve the sparsity of access requests through flow control to satisfy the QoS demands of heterogeneous services \cite{t21,t22}.

According to the diverse communication scenarios with various QoS requirements in terms of delay, connection reliability, data rate, etc., the access priority of the user can be divided into several classes.
Proper access management schemes based on access priority, such as ACB, can mitigate the conflicts of traffic, which has been widely applied in overloaded networks \cite{t11,t23,t24,t25}.
An extended random access scheme provides access links for devices who passed the ACB check, and grant more access links if there are extra resources available \cite{t11}.
Moreover, when the information of resource configuration is available, the optimal ACB factor can be derived \cite{t23}. Fixed ACB control schemes cannot satisfy the requirements of access management in time-varying environments.
A priority-based ACB control scheme divides the users into multiple classes according to latency requirements, and the dynamic adjustment of ACB factor based on priority ensures the QoS of delay-sensitive services \cite{t24}.
An adaptive learning automaton scheme can well excavate potentials from closed-loop interactions with the environment, and realize effective inference and control of traffic \cite{t25}.

The emergence of O-RAN has unlocked the operability of intelligent networks through open interfaces. Specifically, the near-RT RICs in O-RAN have enabled an intelligent access control mechanism to provide customized services for diversified services.
Closed-loop controlling enables xAPPs to track the measurement information in real time and dynamically configure the gNB. Some key issues remain to be addressed when incorporating artificial intelligence in O-RAN,
such as what kind of model should be deployed in a specific scenario, which dataset is suitable for learning, and how to design a solution with the lowest complexity for the real world systems, etc. \cite{t26,t27}.
ML-based automation of O-RAN modules has enabled complex controlling through internal processing and interactions in a black-box model \cite{t28}. In order to further enable the black box model to be applicable in diverse scenarios,
RL methods and agents, i.e., RL operations, can be applied to implement the process of training, testing, and lifecycle verification \cite{t29}. The feasibility and superiority of RL operations have been verified, and it can be easily deployed on O-RAN \cite{t30}.

In this way, the near-RT RIC can dynamically adjust the policies to adapt to the time-varying environment, and make fast and rational decisions based on the feedback information.
It has become a promising solution to make use of the RL method to dynamically determine the strategy of random access that can adapt to diverse scenarios \cite{t31,t32,t33,t34,t35}.
For instance, the inefficiency of traditional access strategies in massive access and the possibility of dynamic access control by Q-learning are demonstrated in \cite{t31}.
An RL-based scheme was proposed for uRLLC services, which can maximize the number of nodes that can be stably served over a long-term duration \cite{t32}.
Some RL-based schemes are devised to adaptively allocate random access resources to services \cite{t33,t34}. A Q-learning enabled scheme with dynamic adaptation of backoff frame was proposed for random access process,
which can mitigate the access conflicts with fixed frame length and improve the network throughput \cite{t35}.

Furthermore, precise and continuous control is required in future data-driven and intelligent O-RAN, where large-scale datasets can be utilized to train an enhanced intelligent agent.
 In this regard, the deep neural networks and deep learning technology popularly applied can be employed. A properly constructed deep neural network can approximate complex nonlinear mapping via training,
  and automatically learn the implicit optimal rules in the closed-loop control strategy. Therefore, the emerging and popular technology of DRL, which combines deep neural networks with RL,
  can be adopted, as has been heavily investigated in literature. For example, a DRL-based multiple access scheme set an objective function based on the total throughput,
  where the network parameters were updated to achieve continuous and optimal selection of the access time slot \cite{t36}. Based on the architecture of deep Q network (DQN),
  an intelligent scheme of node access control was able to avoid frequent switching and achieve better performance of control \cite{t37}. By collecting experiences in higher dimensions,
  a DQN-based scheme can adaptively determine the access contention windows when switching between multiple services \cite{t38}. Besides, for the access control problem in heterogeneous networks,
  the inconsistence in different MAC protocols and  channel capacity can be accommodated by a DRL-based algorithm \cite{t39}.

\vspace{0.1in}
\section{system model}\label{sec:sm}
The massive random access model in the paradigm of O-RAN considered in this paper is illustrated in Fig. 1. Assume that there are in total $N$ potential users within the network.
Let ${\cal U}$ represent the set of all the potential users, and let ${{\cal U}_a}$ represent the set of the active users. An activity indicator ${\alpha _n}$ is used to indicate the activity of the $n$-th user,
i.e., ${\alpha _n}=1$ indicates that the $n$-th user has a demand to access the network at the current time slot, and otherwise it is equal to zero.
In the realistic process of establishing a connection with the gNB, an active user with a single antenna sends a unique preamble to the gNB equipped with $M$ antennas in the radio unit, where the preamble is previously assigned to the active user by the gNB.

\begin{figure}[t]
  \begin{center}
  \vspace{-0.2cm}
  \setlength{\abovecaptionskip}{-0.00cm}
  \setlength{\belowcaptionskip}{-0.2cm}
  \includegraphics[width=3.5 in]{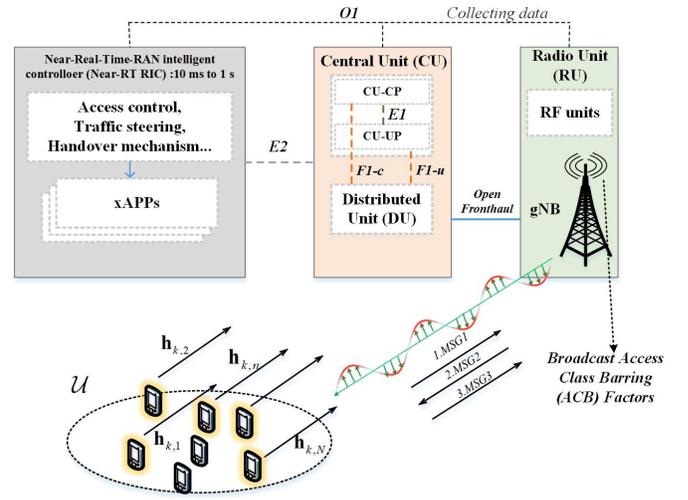}\\
  \caption{Massive random access model in the O-RAN paradigm: The gNB serves as an intelligent agent and physical-layer access point for a massive number of users with different priority ACB classes; Different QoS and access requests required by heterogeneous vertical services, time-varying channel environments, and various scenarios should be dynamically supported with adaptive switching capability.}\label{Fig:algorithm}
  \end{center}
\end{figure}

\begin{figure*}[t]
  \begin{center}
  \vspace{-0.0cm}  
  \setlength{\abovecaptionskip}{-0.10cm}   
  \setlength{\belowcaptionskip}{-0.1cm}   
  \includegraphics[width=5.5 in]{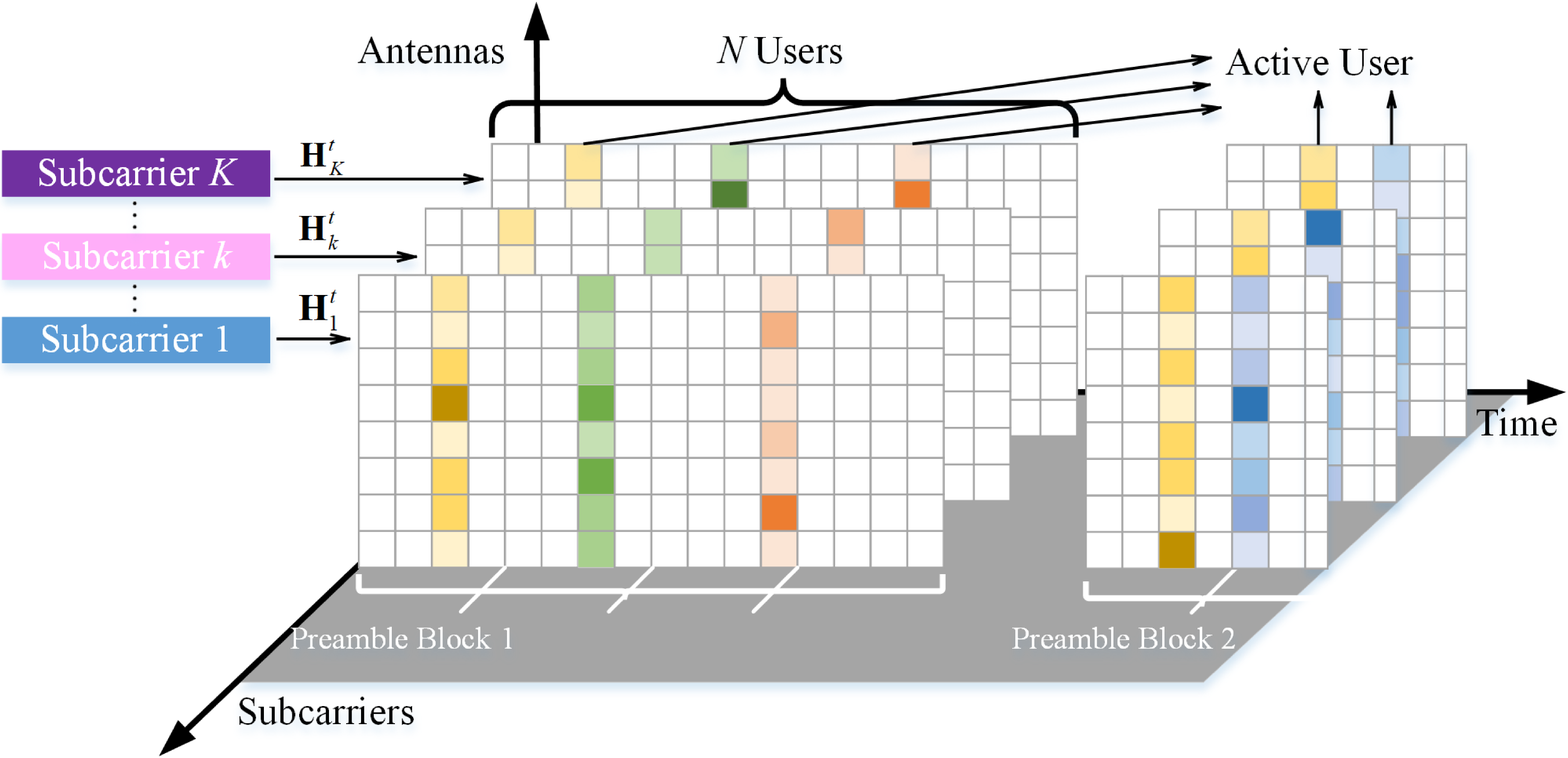}\\
  \caption{Visualization of the massive random access requests, with a series of preamble blocks sent by the active users with access requests represented in the time,
  subcarrier and antenna domains. The size of each preamble block is determined by the number of potential users in the network and the number of antennas at the gNB.
  On a certain time-frequency resource, the subset of active users with access requests shows sparse characteristics compared to the set of the total potential users in the network.}\label{system}
  \end{center}
\end{figure*}

During a typical process of a contention free random access link establishment, three signaling messages denoted by MSG1, MSG2 and MSG3 are sent between the active user and the gNB to convey the unique information of the active user,
the response of the gNB, and the acknowledge character of the active user. To be specific, in time slot $t$, active users modulate their unique preambles ${{\bm{\lambda }}^t} = {[\lambda _1^t,...,\lambda _N^t]^T}$,
i.e., MSG1, onto $K$ OFDM sub-carriers for transmission. The channel matrix of sub-carrier $k$ from $N$ potential users to the gNB is denoted by ${\bf{H}}_k^t = [{\bf{h}}_{k,1}^t,...,{\bf{h}}_{k,N}^t]$,
where ${\bf{h}}_{k,n}^t \in {\mathbb{C}^M}$ denotes the channel response vector from the single-antenna user $n$ to the $M$-antenna gNB. Thus, the measurement vector ${\bf{y}}_k^t \in {\mathbb{C}^M}$ on sub-carrier $k$ received by the gNB is represented as
\begin{equation}\label{sec4:h}\small
  {\bf{y}}_k^t = \sum\limits_{n = 1}^N {\alpha _n^t\lambda _n^t} {\bf{h}}_{k,n}^t + {{\bf{z}}^t} = \underbrace {{\bf{H}}_k^t{{\bm{\Lambda }}^t}}_{{\bf{\tilde H}}_k^t}{\bm{\alpha }}_{}^t + {{\bf{z}}^t},
\end{equation}
where ${{\bm{\Lambda }}^t} = {\rm{diag}}\{ \lambda _1^t,...,\lambda _N^t\}$ is a diagonal matrix whose diagonal elements are composed of the preamble vector ${{\bm{\lambda }}^t}$,
and ${\bf{z}}^t \in {\mathbb{C}^{M}}$ represents the background noise. For simplicity of notation, let ${\bf{\tilde H}}_k^t$ represent the normalized channel matrix ${\bf{H}}_k^t{{\bm{\Lambda }}^t}$,
which represents the original channel matrix ${\bf{H}}_k^t$ normalized by the preambles ${{\bm{\Lambda }}^t}$. It is worth noting that, the normalized channel matrix ${\bf{\tilde H}}_k^t$ can be regarded as an observation matrix in the framework of CS.
If the unknown activity indicator vector ${{\bm{\alpha }}^t} = {[\alpha _1^t,...,\alpha _N^t]^T}$ in (1) has a sparse property, i.e., the number of active users are much less than the total users, and the observation matrix satisfies the restricted isometry property,
it is high probable that it can be recovered with bounded error from the measurement vector ${\bf{y}}_k^t$ according to the CS theory \cite{t5}. Then, the active users are detected by recovering the activity indicator vector ${{\bm{\alpha }}^t}$ accurately.

The mechanism of massive random access requests with a series of preamble blocks sent by the active users with access requests represented in the time, subcarrier and antenna domains is visualized as shown in Fig. 2.
Assuming that the access requests of the active users are sparse with respect to all the potential users within the network, then the SAUD problem can be modeled as a sparse recovery problem to estimate the support, i.e.,
the positions of nonzero entries, of the unknown activity indicator vector ${{\bm{\alpha }}^t}$. To this end, the sparse recovery problem to estimate ${{\bm{\alpha }}^t}$ can be decomposed into $K$ sub-problems for each sub-carrier as given by (1).
Specifically, a sub-carrier-wise activity indicator vector, i.e., $\widehat {\bm{\alpha }}_k^t = {[\hat \alpha _{k,1}^t,...,\hat \alpha _{k,N}^t]^T}$, can be estimated first by solving the sub-problem ${\bf{y}}_k^t = {\bf{\tilde H}}_k^t{\bm{\alpha }}_k^t + {{\bf{z}}^t}$ for each sub-carrier $k = 1 \cdots K$.
Then, the estimated sub-carrier-wise activity indicator vector $\widehat {\bm{\alpha }}_k^t$ can be regarded as a voter to vote for the nonzero positions of the final estimated activity indicator vector $\widehat {\bm{\alpha }}_{}^t$:
The $n$-th element $\hat \alpha _n^t$ of $\widehat {\bm{\alpha }}_{}^t$ will be marked as one if the number of votes for position $n$ exceeds a certain decision threshold ${\upsilon _{{\rm{th}}}}$ and marked as zero otherwise, which is given by
\begin{equation}\label{sec4:h}\small
  \hat \alpha _n^t = \left\{ {\begin{array}{*{20}{l}}
  {1,}&{\frac{1}{K}\sum\limits_{k = 1}^K {\hat \alpha _{k,n}^t} {\rm{ > }}{\upsilon _{{\rm{th}}}}{\rm{,}}}\\
  {0,}&{{\rm{ otherwise}}{\rm{.  }}}
  \end{array}} \right.
\end{equation}
After the active users are detected by SAUD via (2), the gNB sends a random access response, i.e., MSG2, to the detected users.

Starting from sending their preambles, the active users start a timer window to capture the feedback of gNB.
If a user successfully parses the response corresponding to the previous MSG1 within this timer window, the device will feed back an acknowledge character (ACK),
i.e., MSG3, to the gNB, indicating that the connection has been successfully established. Otherwise, it is considered as reception failure.

\vspace{0.1in}
\section{Reinforcement-Learning-assisted Sparse Active User Detection for Access Control in O-RAN}\label{sec:sm}
The performance of SAUD relies heavily on the sparsity of access requests. Thus, the access flow control of the potential users in the network should be carefully and properly managed.
To this end, we propose an RL-assisted SAUD (RL-SAUD) scheme in this section, which introduces the ACB mechanism to coordinate the access requests of the users.
The scheme aims to adjust the ACB factor reasonably and dynamically, and improve the detection accuracy of the SAUD while allowing as many users as possible to access to satisfy the requirements of various services.
We will first introduce the ACB mechanism for flow control in O-RAN context, and then present the proposed RL-SAUD scheme in detail in this section.

\subsection{Access Class Barring (ACB) for Flow Control}
It should be noted that, the accuracy of the SAUD problem as given by (1) and (2) is closely related with the sparsity level of the access requests.
In various heterogeneous services of O-RAN with excessively massive concurrent access requests, the sparsity of the activity indicator vector in problem (1) may be destroyed,
which reduces the probability of accurate user detection and results in access failure. Therefore, the gNB needs a reasonable strategy to control the access traffic,
and ACB is a good candidate to achieve this goal.

In fact, ACB is a flow control mechanism that divides the users into multiple classes according to their access priorities. Suppose there are $L$ classes, the users in class $l$ are represented by the set ${{\cal U}_l}$,
and the number of elements in this set is ${N_l}$. The near-RT RIC generates different ACB factors $\{ {p_l} \in [0,1]\} _{l = 1}^L$ for each of the $L$ classes to perform traffic management via a procedure called ACB check.
Specifically, in the process of ACB check, a certain active user $n \in {{\cal U}_l}$ with access requirements randomly samples a value ${q_n} \in [0,1]$ before sending its preamble:
Only if ${q_n} \le {p_l}$ will user $n$ send the preamble, otherwise it will back off to a random sampling time within a predefined range and wait for the next ACB check procedure \cite{t40}.

In the context of O-RAN, the pre-trained models of various heterogeneous services are stored in the xApps directory, which provides convenience for intelligent ACB regulation.
When an active user initiate an access request, the near-RT RIC instantiates its xAPP in the directory and controls the underlying centralized unit and distributed unit through the E2 interface.
Finally, the gNB broadcasts the ACB configuration information to control the traffic in the network.
With the support of the ACB flow control mechanism, the number of devices accessing the network in the same time period can be properly controlled and coordinated,
which helps preserve the sparsity of the access requests of active users, i.e., the sparsity of the activity indicator vector ${{\bm{\alpha }}^t}$ as given in (1).

According to related research,
the accuracy of SAUD with respect to the number of access requests is a monotonically decreasing convex curve \cite{t24}. When flow control is not performed in case of massive concurrent
access requests, i.e., when the ACB factor is set to one, all the active users can pass the ACB check procedure and get access to the network, which makes the activity
indicator vector ${{\bm{\alpha }}^t}$ no longer sparse and results in severe conflicts. In this case, it is difficult or impossible to accurately detect and distinguish between the active users from
the received signal in (1). Hence, a properly designed mechanism of flow control of the active users should be investigated to relieve the conflicts and preserve the sparsity of
access activity, so as to support massive access and the coexistence of various heterogeneous services in the complex and time-varying environments in O-RAN. To this end,
a closed-loop control scheme based on RL is devised to adjust the ACB factors in real time, which is then broadcast to all the users within the network,
as described in detail in the next sub-section.

\subsection{Reinforcement Learning-Assisted Sparse Active User Detection (RL-SAUD) for Massive Random Access}
First, we present the model of interactions between the intelligent agent, i.e., the gNB, and the environment, i.e., the users within the RAN, which can be regarded as a Markov decision process (MDP) as illustrated in Fig. 1.
In the context of O-RAN, the closed-loop control is actually an MDP process, in which the xAPP plays the role of an intelligent agent enabled by the RL framework, and the near-RT RIC serves as a container of various xAPPs.
The details of the interactions in the closed-loop access control process are listed as follows.
\begin{itemize}
  \item {\color{black}The near-RT RIC throws an ACB vector ${{\bf{p}}^t}$ through the open fronthaul interface, and the gNB broadcasts it to the users with different access classes in the RAN. Given the ACB vector ${{\bf{p}}^t}$, the number of access-permitted users can be approximately estimated by aggregating over all the $L$ classes, i.e., $N_{\rm{p}}^t = \sum\limits_{l = 1}^L {p_l^t} N_l^t$.}
  \item The active users that have passed the ACB check send a unique preamble to the gNB. Then the gNB starts to detect the identification of the users via the SAUD algorithm, and sends an MSG2 to the detected users. Once the correctly detected user receives the MSG2, an MSG3, i.e., ACK, is returned to the gNB.
  \item The gNB estimates the approximate number of valid access users $N_{\rm{v}}^t$ by counting the number of feedback ACKs and calculates the user detection accuracy $c^t$ in the current time slot $t$.
  \item The near-RT RIC extracts the statistical data of the current environment through the E2 interface as shown in Fig. 1, updates the policy of traffic management in the xAPP, and performs the ACB control action determined by the policy for the next time slot.
\end{itemize}

The proposed RL-assisted SAUD algorithm is summarized in {\bf{Algorithm 1}}. Specifically, to apply the Q-learning method, the ACB factor for class $l$ is quantized into discrete $X_1$ levels, i.e., $p_l^t \in \Omega  \buildrel \Delta \over = \left\{ {i/{X_1},1 \le i \le {X_1}} \right\}$,
where $\Omega$ is the set of feasible actions, i.e., ACB factor values, for a certain class of users. After the selected action is performed, i.e., broadcast to all the users,
different classes of users perform the ACB check procedure, which generates a set of active users ${{\cal U}_a}$ and a corresponding activity indicator vector ${{\bm{\alpha }}^t}$.
Then, the active users in the set ${{\cal U}_a}$ will establish a connection with the gNB. The gNB estimates the activity indicator vector ${\widehat {\bm{\alpha }}^t}$ via the SAUD algorithm,
and then transmits MSG2 to the detected active users. The SAUD algorithm is summarized in {\bf{Algorithm 2}}, where ${[{{\bf{\tilde H}}_k}]^*}$ and ${[{{\bf{\tilde H}}_k}]^\dag }$ represent the conjugate
and Moore-Penrose inverse of the normalized channel matrix ${{\bf{\tilde H}}_k}$ of sub-carrier $k$ as given in (1), respectively.

The user who successfully receives MSG2 in the timer window returns an ACK to the gNB, and the gNB estimates the detection accuracy accordingly.
For the convenience of referring to the past information, the \emph{state} of the RL-based algorithm is composed of the ACB factor vector and the detection accuracy at the previous time slot,
i.e., ${{\bf{s}}^t} = \left[ {{{\bf{p}}^{t - 1}},{c^{t - 1}}} \right]$. Each state-action pair has a Q-value, forming a Q-table of size ${X_1}^L{X_2} \times {X_1}^L$,
where $X_2$ is the number of quantization levels of the detection accuracy $c$.

The agent tends to choose the currently optimal action, i.e., ${{\bf{p}}^t} = {{\bf{p}}^*}$ that maximizes the reward value $Q({\bf{s}},{\bf{p}})$ in the current state ${{\bf{s}}^t}$,
but in this way some actions might not be explored, which might result in stuck in a local optimum. In this regard, the $\epsilon$-greedy method provides a certain small probability to adopt a random strategy by setting an $\epsilon $ value,
so that every feasible action might be explored, which is given by
\begin{equation}\label{sec4:h}\small
  \Pr \left( {{{\bf{p}}^t} = {{\bf{p}}^*}} \right) = \left\{ {\begin{array}{*{20}{l}}
    {1 - \epsilon ,}&{{{\bf{p}}^*} = \mathop {\arg \max }\limits_{{\bf{p}} \in {\Omega ^L}} Q\left( {{{\bf{s}}^t},{\bf{p}}} \right)},\\
    {\frac{\epsilon }{{|{X_1} + 1|}},}&{{\rm{ otherwise }}}.
    \end{array}} \right.
\end{equation}
where ${\Omega ^L}$ is the action space. Based on (3), the probability of choosing the action with the largest Q-value is $1-\epsilon $.
\begin{algorithm}[h]
          \caption{Reinforcement Learning assisted Sparse Active User Detection with Traffic Flow Control (RL-SAUD)}
          \textbf{Initialization:}\\
          Q-values $Q({{\bf{s}}^t},{{\bf{p}}^t})$ in Q-table\\
          Initialize a random state ${{\bf{s}}^0}$ \\
          \For{$t = 1, 2, 3,...$}{
            In state ${{\bf{s}}^t}$, choose action ${{\bf{p}}^t}$ via (3)\\
            Calculate $N_{\rm{p}}^t$\\
            Active users to perform ACB check using ${{\bf{p}}^t}$ and send MSG1 to gNB if passed\\
            gNB performs SAUD in {\bf{Algorithm 2}} and feed back MSG2\\
            Active users detect MSG2 and send MSG3 back to gNB\\
            gNB counts the number of valid access users $N_{\rm{v}}^t$ via ACKs in MSG3\\
            Calculate detection accuracy ${c^t} = N_{\rm{v}}^t/N_{\rm{p}}^t$\\
            Formulate the next state ${{\bf{s}}^{t + 1}} = [{{\bf{p}}^t},{c^t}]$\\
            Obtain current system utility $u^t$ via (4)\\
            Update the Q-table using (5)\\
          }
\end{algorithm}
\begin{algorithm}[h]
          \caption{Sparse Active User Detection (SAUD)}
          \textbf{Input:}\\
          Channel matrix ${\{ {{\bf{\tilde H}}_k}\} _{1 \le k \le K}}$\\
          Received signal ${\{ {{\bf{y}}_k}\} _{1 \le k \le K}}$ and step size $s$ \\
          \For{$k = 1, 2, 3,...$}{
            \textbf{Initialization:} \\
            ${F_0} = \emptyset $, $\theta  = 1 $, $i = 1 $ and residual ${{\bf{r}}_0} = {{\bf{y}}_k}$\\
            \textbf{repeat:}\\
            ${{\cal S}_i} = \rm{Max}\left( {|{{[{{{\bf{\tilde H}}}_k}]}^*}{{\bf{r}}_{i - 1}}|,s \times \theta } \right)$\\
            ${{\cal C}_i} = {F_{i - 1}} \cup {{\cal S}_i}$\\
            $F = \rm{Max}\left( {|[{{{\bf{\tilde H}}}_k}]_{{{\cal C}_i}}^\dag {{\bf{y}}_k}|,s \times \theta } \right)$\\
            ${\bf{r}} = {{\bf{y}}_k} - {[{{\bf{\tilde H}}_k}]_F}[{{\bf{\tilde H}}_k}]_F^\dag {{\bf{y}}_k}$\\
            \uIf {terminal condition is met} {
              \textbf{Continue}
            }
            \uElseIf{${\left\| {\bf{r}} \right\|_2} > {\left\| {{{\bf{r}}_{i - 1}}} \right\|_2}$} {
              $\theta  = \theta+1 $
            }
            \uElse {
              ${F_i} = F$, ${{\bf{r}}_i} = {\bf{r}}$, $i=i+1$
            }
            ${\widehat {\bm{\alpha }}_k} = [{{\bf{\tilde H}}_k}]_F^\dag {{\bf{y}}_k}$\\
          }
          Detect the active user $\widehat {\bm{\alpha }}$ via (2)\\
          \textbf{Return:} $\widehat {\bm{\alpha }}$
  \end{algorithm}

The system utility function $u^t$ is designed so as to strengthen the policy of choosing a favorable action over the iterative learning process, which is given by
\begin{equation}\label{sec4:h}\small
  {u^t} = {c^t}\sum\limits_{l = 1}^L {p_l^t{r_l}N_l^t}  - {\rho _1}\frac{1}{L}{\sum\limits_{l = 1}^L {(p_l^t - \overline {{p^t}} )} ^2} - {\rho _2}(1 - {c^t}),
\end{equation}
where $r_l$ is the access priority score for class $l$, with a higher score indicating a higher access priority. The first term to the right of equation (4) represents the quantity
of the valid accessed users weighted by the access priority scores. Intuitively, if more users with higher access priority scores are permitted to access and accurately detected,
i.e., valid accessed, the system utility should get a raise. The second and third terms in (4) both play the role of penalty on the utility.
Specifically, the second term in (4) represents the variance of the elements in ${{\bf{p}}^t}$ weighted by a coefficient $\rho _1$, which plays the role of a penalty on the policy
ignoring the access of the users in low-priority classes. If the variance is large, it implies that the users in some of the low-priority classes are hardly permitted to access the
network, which is \emph{not} a favorable decision especially for mMTC services with massive users of different priority scores required to access. Thus, the agent can guide the algorithm
to learn a policy favorable for massive access control in mMTC services by setting a positive value of the coefficient $\rho _1$ in the utility function (4).

On the other hand, for uRLLC services, the reliability and stability are utmost important. In this case, the third term in (4) plays the role of a penalty on the detection error
weighted by a coefficient $\rho _2$. Thus, the agent can easily switch to a policy favorable for accurate and reliable detection in uRLLC services simply by setting a positive value
of $\rho _2$ to include penalty on detection error. Consequently, properly setting the two coefficients $\rho _1$ and $\rho _2$ for the two penalty terms will lead to an
appropriate tradeoff between different QoS requirements of various heterogeneous services, and provide good support of flexible switching between them.
\begin{figure*}[t]
  \begin{center}
  \vspace{-0.0cm}  
  \setlength{\abovecaptionskip}{-0.10cm}   
  \setlength{\belowcaptionskip}{-0.1cm}   
  \includegraphics[width=6 in]{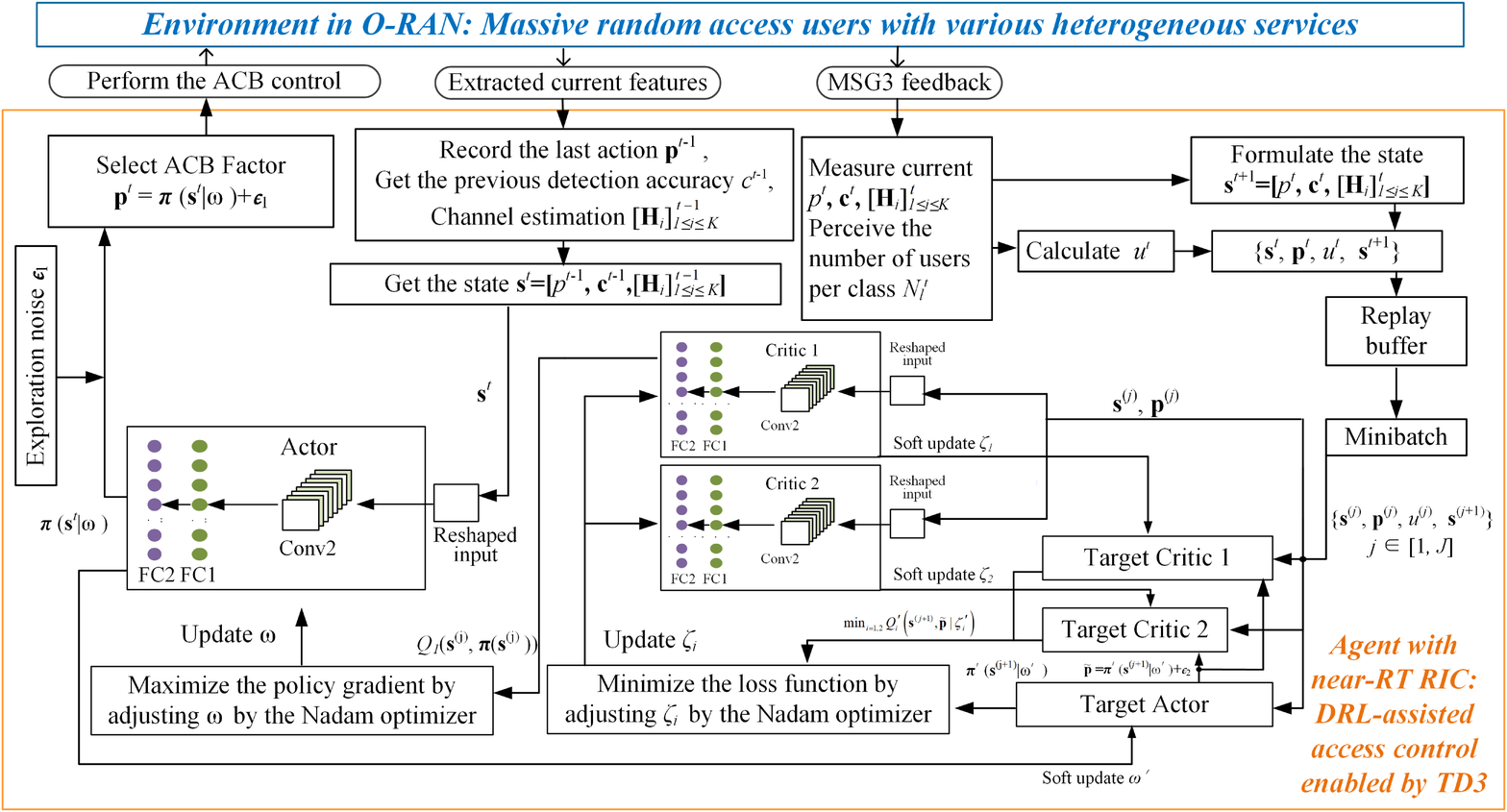}\\
  \caption{Architecture of TD3-enabled dynamic control of ACB factors: Active users get access via the ACB factors determined by the intelligent agent at the gNB, powered by the DRL-assisted access control policy in the near-RT RIC.}\label{system}
  \end{center}
\end{figure*}

The values of the Q-table are updated after each time slot $t$ using the Bellman equation as given by
\begin{equation}\label{sec4:h}\small
  Q({{\bf{s}}^t},{{\bf{p}}^t}) \leftarrow (1 - \varpi )Q({{\bf{s}}^t},{{\bf{p}}^t}) + \alpha ({u^t} + \beta {\max _{\widehat {\bf{p}} \in {\Omega ^L}}}Q({{\bf{s}}^{t + 1}},\widehat {\bf{p}})),
\end{equation}
where $\varpi$ and $\beta $ represent the learning rate and the discount rate, respectively. The update of the Q-table is driven by the system utility function $u^t$ as given in (4).

It is worth noting that, the Q-learning method adopted by the proposed RL-SAUD scheme is a discrete action control approach, which is equivalent to sampling from the exact policy.
Thus, the quantization interval of the states and actions should be smaller in order to convey more specific information to achieve satisfactory performance \cite{t41}.
However, this results in an exponential increase in the size of the Q-table, which costs too much computational complexity. In addition, each interaction is only learned once and
the experience is not well exploited for future learning, which limits the potentials of big data-driven approaches. Therefore, it is necessary to introduce the DRL technique,
i.e., a data-driven paradigm enabled by training the deep neural networks, to resolve the performance limitation due to discrete quantification and dimensional curse of continuous
state and action spaces, and make full use of previous experiences for faster convergence towards learning the optimal strategy.

\vspace{0.1in}
\section{Deep-Reinforcement-Learning-Assisted Sparse Active User Detection for Access Control in O-RAN}
In this section, a twin delayed deep deterministic (TD3) policy gradient algorithm based on the Actor-Critic framework is introduced to dynamically adjust the ACB factor and properly coordinate the access requests,
thereby improving the number of valid access users and the accuracy of SAUD. Compared with the RL-SAUD scheme, the proposed DRL-SAUD scheme uses data-driven approaches to train deep neural networks to resolve the problem
of quantization loss and high-dimensional state and action spaces, and accelerate the convergence rate of the access control strategy. In fact, the TD3 algorithm adopted in the DRL-SAUD scheme is a cutting-edge
DRL-based algorithm with some favorable features: i) The Actor-Critic framework underlying TD3 is very helpful for offline model testing; ii) Deep neural networks enable the state to refer to more complex
indicators such as channel state information; iii) Better to realize continuous action control, and iv) Accelerating the convergence of learning via experience replay.

Different from the RL-based Q-learning algorithm that selects the action with the largest Q-value in the Q-table at a certain state, the DRL-based TD3 approach utilized in this
paper employs the Actor-Critic framework, which is illustrated in Fig. 3. Specifically, an evaluation network, i.e., the Critic, processes and learns the experience obtained by
the policy network, i.e., the Actor, and then passes the Q-value to the Actor for learning. In this way, the Actor is responsible for action decisions, and the Critic is
responsible for scoring the actions. In the framework of TD3, closed-loop offline training can be performed using the Actor-Critic networks to optimize for an effective model before it is applied to the realistic environment,
which can significantly improve the testing performance of the model and the user experience compared with the purely online learning method, especially at the beginning of the testing phase \cite{t43}.

The devised architecture of TD3-enabled dynamic control of ACB factors is as shown in Fig. 3. Specifically, the TD3 architecture consists of six networks in total,
including the current Critic 1, Critic 2 and Actor, and their corresponding target networks. The current networks are intended to interact with the environment in real time,
and the target networks are responsible for providing reference values for updating the current networks. By employing two Critic networks, the agent can choose a smaller
Q-value during the update process to avoid overestimation of Q-values \cite{t42}.

The proposed DRL-SAUD algorithm enabled by TD3 is summarized in detail in {\bf{Algorithm 3}}. Different from the RL-based scheme, the state ${{\bf{s}}^t} = \left[ {{{\bf{p}}^{t - 1}},{c^{t - 1}},\left[ {{{\bf{H}}_k}} \right]_{1 \le k \le K}^{t - 1}} \right]$ is
formulated by directly concatenating the real continuous-valued action ${{\bf{p}}^{t - 1}}$,  detection accuracy ${c^{t - 1}}$, and channel matrices $\left[ {{{\bf{H}}_k}} \right]_{1 \le k \le K}^{t - 1}$ at the previous
time slot without quantization. The high-dimensional features of the state can be extracted by the deep neural networks in the Actor, and then a continuous action can be determined and output via the policy $\pi ({\bf{s}}|\omega )$.
Random exploration of the agent is achieved by adding an additive noise term with variance of ${\epsilon _1}\sim{\cal N}(0,\sigma )$ instead of the $\epsilon$-greedy method,
so that the final action to be performed is slightly modified as given by ${{\bf{p}}^t} = \pi ({{\bf{s}}^t}|\omega ) + {\epsilon _1}$.
It should be noted that a too small value of the additive noise will make the action no longer exploratory, while a too large value will refrain the exploitation of the learnt policy,
so a proper tradeoff between exploration and exploitation can be achieved by setting an appropriate value of the additive noise.

After the ACB check, the active users start requesting access to the gNB, and then the gNB obtains the next state ${{\bf{s}}^{t + 1}} = \left[ {{{\bf{p}}^t},{c^t},\left[ {{{\bf{H}}_k}} \right]_{1 \le k \le K}^t} \right]$ and $u^t$.
The information including the current and next states along with the current utility and action will be packed into a transition as an experience, i.e., ${\Im ^t} = \{ {{\bf{s}}^t},{{\bf{p}}^t},{u^t},{{\bf{s}}^{t + 1}}\}$,
and stored in an experience replay buffer ${\cal B}$. The replay buffer enables previous experiences to be learned from repeatedly for faster convergence to the optimal model.
Meanwhile, outdated transitions will be replaced by the latest ones to keep track of the variation of the environment.

When a number of transitions have been captured in the replay buffer, the agent randomly selects ${\cal J}$ transitions from the replay buffer,
i.e., $\{ {{\bf{s}}^{(j)}},{{\bf{p}}^{(j)}},{u^{(j)}},{{\bf{s}}^{(j + 1)}}\} ,j \in [1,{\cal J}]$ to update the weights of the networks in real time.
The Target Actor first computes a reference action $\widetilde {\bf{p}}$ for ${{\bf{s}}^{(j + 1)}}$
\begin{equation}\label{sec4:h}\small
  \widetilde {\bf{p}} \leftarrow \pi '({{\bf{s}}^{(j + 1)}}|\omega ') + {\epsilon _2},\\
  {\epsilon _2}\sim{\mathop{\rm clip}\nolimits} ({\cal N}(0,\tilde \sigma ), - g,g),
\end{equation}
where the policy noise ${\epsilon _2}$ is a random variable following a normal distribution clipped by $\pm g$. The additive policy noise ${\epsilon _2}$ can smooth the Q-function as the output of the Critic network, and enhance the reliability of the Q-value provided by the Critic with the fluctuation of the actions.
Then a reference value ${y_{\rm{r}}}$ of the Q-value ${\left\{ {{Q_i}({{\bf{s}}^{(j)}},{{\bf{p}}^{(j)}})} \right\}_{i = 1,2}}$ is given by
\begin{equation}\label{sec4:h}\small
  y_{\rm{r}}^{(j)} \leftarrow {u^{(j)}} + \gamma {\min _{i = 1,2}}{Q_i}^\prime \left( {{{\bf{s}}^{(j + 1)}},\widetilde {\bf{p}}|{\zeta _i}^\prime } \right),
\end{equation}
where $\gamma $ is a discount factor. The Critic network uses the Nadam optimizer to minimize the loss function as given by
\begin{equation}\label{sec4:h}\small
  {\zeta _i} \leftarrow \mathop {\arg \min }\limits_\zeta  \frac{1}{{\cal J}}\sum\limits_j {{{\left( {y_{\rm{r}}^{(j)} - {Q_i}({{\bf{s}}^{(j)}},{{\bf{p}}^{(j)}}|\zeta )} \right)}^2}} .
\end{equation}
Different from the update of the RL-based scheme, the DRL-based scheme manipulates more data in one epoch of training, and a newly recorded experience can be learned several times in subsequent training.

The policy network is updated more slowly than the evaluation network, which ensures that the Critic has minimized its own estimation error before providing scores for policy updates \cite{t43}.
In the design of the proposed scheme in this paper, the Critic is updated $d$ times every time the Actor is updated, and the Nadam optimizer is adopted to maximize the policy gradient as given by
\begin{equation}\label{sec4:h}\small
  \omega  \leftarrow \mathop {\arg \max }\limits_\omega  \frac{1}{{\cal J}}{\nabla _{\bf{p}}}{Q_1}({\bf{s}},{\bf{p}}|{\zeta _1}){|_{{\bf{s}} = {{\bf{s}}^{(j)}},{\bf{p}} = \pi ({{\bf{s}}^{(j)}})}}{\nabla _\omega }\pi ({\bf{s}}|\omega ){|_{{\bf{s}} = {{\bf{s}}^{(j)}}}}.
\end{equation}
In (9), ${\nabla _{\bf{p}}}{Q_1}({\bf{s}},{\bf{p}}|{\zeta _1})$ represents the gradient of ${Q_1}({\bf{s}},{\bf{p}}|{\zeta _1})$ with respect to ${\bf{p}}$,
and ${\nabla _\omega }\pi ({\bf{s}}|\omega )$ is the gradient of $\pi ({\bf{s}}|\omega )$ with respect to $\omega$.

Every time the Actor is updated, a soft update is also performed on each of the target networks, which is given by
\begin{equation}\label{sec4:h}\small
  {\zeta _i}^\prime  = \delta {\zeta _i} + (1 - \delta ){\zeta _i}^\prime, \omega ' = \delta \omega  + (1 - \delta )\omega ',
\end{equation}
where $\delta  \in (0,1]$ is a memory coefficient that can be properly set to achieve tradeoff between convergence rate and accuracy.

\begin{algorithm}[h]
          \caption{DRL-Assisted Sparse Active User Detection Enabled by TD3 for Access Flow Control (DRL-SAUD)}
          \textbf{Initialization:}\\
          Actor network $\pi ({\bf{s}}|\omega )$\\
          Critic1 network ${Q_1}({\bf{s}},{\bf{p}}|{\zeta _1})$, Critic2 network ${Q_2}({\bf{s}},{\bf{p}}|{\zeta _2})$\\
          Target network $\pi '({\bf{s}}|\omega ')$, ${Q_1}^\prime ({\bf{s}},{\bf{p}}|{\zeta _1}^\prime )$, ${Q_2}^\prime ({\bf{s}},{\bf{p}}|{\zeta _2}^\prime )$\\
          Reset replay buffer ${\cal B}$ \\
          Generate a random initial state ${{\bf{s}}^0}$ \\
          \For{$t = 1, 2, 3,...$}{
            Determine the ACB factor action ${{\bf{p}}^t} = \pi ({{\bf{s}}^t}|\omega ) + {\epsilon _1}$ for state ${{\bf{s}}^t}$\\
            Active users send MSG1\\
            gNB performs SAUD in {\bf{Algorithm 2}} to detect active users, and then feeds back MSG2 \\
            Detected active users receives MSG2 and returns ACK\\
            gNB formulates next state ${{\bf{s}}^{t + 1}} = [ {{{\bf{p}}^t},{c^t},\left[ {{{\bf{H}}_k}} \right]_{1 \le k \le K}^t}]$ and derive current system utility $u^t$\\
            Store transition $\{ {{\bf{s}}^{(j)}},{{\bf{p}}^{(j)}},{u^{(j)}},{{\bf{s}}^{(j + 1)}}\}$ in replay buffer ${\cal B}$\\
            Randomly sample $\cal J$ transitions from ${\cal B}$ \\
            Obtain the reference action $\widetilde {\bf{p}} \leftarrow \pi '({{\bf{s}}^{(j + 1)}}|\omega ') + {\epsilon _2}$\\
            Calculate the reference Q-value ${y_{\rm{r}}}$ using (7)\\
            Update ${\zeta _i}$ via (8)\\
            \If {($t$ mod $d=0$)} {
              Update $\omega$ via (9)\\
              Soft update weights of target networks ${\zeta _i}^\prime$ and $\omega '$ using (10)\\
            }
            $t=t+1 $
            }
\end{algorithm}

\vspace{0.1in}
\section{Performance Analysis and Evaluation}
In this section, we present theoretical performance analysis and evaluation on some important issues related with the proposed schemes.
First, for the SAUD algorithm, the detection accuracy with respect to the sparsity of the user access requests is theoretically analyzed.
Then, for the RL-based schemes, the theoretical bound of the utility function is derived, and the computational complexity of the proposed schemes is evaluated.

\subsection{Performance Analysis of SAUD}
The SAUD problem as formulated in (1) can be convex relaxed by ${\ell _1}$-norm minimization of the unknown sparse activity indicator vector \cite{t44}, which yields a convex optimization problem as given by
\begin{equation}\label{sec4:h}\small
  {\bm{\hat \alpha }} = \mathop {\arg \min }\limits_{{\bm{\alpha }} \in {\mathbb{C}^{N}}} \left\{ {\frac{1}{{2M}}\left\| {{\bf{y}} - {\bf{\tilde H\alpha }}} \right\|_2^2 + {\varepsilon _N}{{\left\| {\bm{\alpha }} \right\|}_1}} \right\},
\end{equation}
where the first term in the minimization problem represents the regularization of the ${\ell _2}$-norm error of sparse recovery. The second term is the ${\ell _1}$-norm of the unknown sparse activity indicator vector ${\bm{\alpha }}$,
which encourages a sparse solution of ${\bm{\alpha }}$. A coefficient ${\varepsilon _N}$ is adopted to make tradeoff between the measurement error due to sparse recovery and the
sparsity requirement of the active users, which is given by ${\varepsilon _N} = \sqrt {\frac{{2\varphi \log N}}{{\eta N}}} ,\varphi  > 2$. For example, when the number of user access
requests increases sharply, the value of ${\varepsilon _N}$ can be reduced to moderately relax the requirement of access sparsity. After performing SAUD,
the set of active users can be obtained from the support of the recovered activity indicator vector ${\bm{\hat \alpha }}$.
The active user detection accuracy $c$ is then obtained by the ${\ell _1}$-norm difference between the recovered and the real activity indicator vectors, which is given by
\begin{equation}\label{sec4:h}\small
  c = 1 - \frac{{{{\left\| {{\bm{\alpha }} - {\bm{\hat \alpha }}} \right\|}_1}}}{N}.
\end{equation}

\begin{theorem}
  The active user detection accuracy $c$ is lower-bounded by
  \begin{align}\label{align:w*}
    c \ge 1 - {a_1}\exp \left( { - {a_2}{\rm{ min}}\left\{ {{{\left\| {\bm{\alpha }} \right\|}_1},{\rm{ log}}(N - {{\left\| {\bm{\alpha }} \right\|}_1})} \right\}} \right),
  \end{align}
if the following conditions are satisfied
\begin{align}\label{align:w0cons}\small
    \begin{split}
      &{{{\left\| {\bm{\alpha }} \right\|}_{1,\max }} \ge \frac{{\eta N(1 - \frac{1}{\varphi })}}{{2\log N}}}\\
      &{{{\left| {{\lambda _n}} \right|}_{\min }} > \vartheta {\varepsilon _N} = \vartheta \sqrt {\frac{{2\varphi \log N}}{{\eta N}}} .}
    \end{split}
\end{align}
\end{theorem}
\indent\indent \emph{Proof.} See the details in Appendix A.\QEDA

$\textbf{Remark 1}$. It is indicated from {\bf{Theorem 1}} that, the active user detection accuracy is closely related with the sparsity of the user access requests.
  The two constraints in (14) are the constraint on the sparsity of the user access requests and the constraint on the minimum amplitude of the user preamble ${\left| {{\lambda _n}} \right|_{\min }}$,
  respectively. ${\left\| {\bm{\alpha }} \right\|_{1,\max }}$ denotes the maximum sparsity level of the activity indicator vector that is tolerable in the network, which is subject to the constraint as given by
  \begin{align}\label{align:w*}
    M = 2\left( {{{\left\| {\bm{\alpha }} \right\|}_{1,\max }} + \frac{1}{{\varepsilon _N^2}}} \right)\log \left( {N - {{\left\| {\bm{\alpha }} \right\|}_{1,\max }}} \right).
  \end{align}

  For the sparse measurement model as given in (1), $\eta  = M/N$ represents the compressive measurement ratio, which is the ratio of the measurement data size, i.e.,
  the number of antennas, to the length of the sparse vector, i.e., the total number of potential users. The parameters of $\vartheta$ and $\varphi $ are set manually in the optimization process.

\subsection{Theoretical Bound and Computational Complexity of Reinforcement Learning Based Schemes}
To evaluate the theoretical performance of the proposed RL-based and DRL-based schemes, the system utility as in (4) can be derived in closed-form for a typical case with each access priority score identical, which is presented in the following theorem.
\begin{theorem}
  If the access priority score of each user class is identical, the problem of maximizing the system utility function $u$ as given in (4)
  is turned into a convex problem that has a tractable theoretical bound.
\end{theorem}
\indent\indent \emph{Proof.} See the details in Appendix B.\QEDA

$\textbf{Remark 2}$. When there is no difference in access priority of the users in the network, i.e., $L=1$, the first penalty term in (4) intended to mitigate the variance of the ACB factors of different priority classes will disappear.
  In this case, the RL-based and DRL-based schemes are actually dynamically searching for a proper policy of access control to approximate the optimal solution or a sub-optimal solution towards the theoretical bound derived in {\bf{Theorem 2}}.

  When different user classes have different access priority scores, more uncertainty is brought to the network.
  Considering the time-varying property of the environment and the diverse channel conditions and QoS requirements of different user classes,
  it is not guaranteed that the optimal solution of the original problem is still tractable in closed-form.
  Hence, the ability of RL-based schemes in searching for a sub-optimal solution towards the system utility in complex environments over a reasonable time frame can be exploited.
  In the decision process of the RL-based and DRL-based schemes, there are two tradeoffs that need to be considered:
  The tradeoff between the number of high-priority permitted-access users and the variance in the number of permitted-access users with different priority scores,
  and the tradeoff between the total number of permitted-access users and the active user detection accuracy.

Next, we will investigate the computational complexity of the proposed RL-based and DRL-based schemes. According to the related research in literature \cite{t45},
when the number of training episodes of an RL-based algorithm is $\tau $ with each episode including $\xi$ time slots, the computational complexity is given by
\begin{equation}\label{sec4:h}\small
  {{\cal T}_1} = {\cal O}(\tau \xi )
\end{equation}
if $\tau \xi  > {\rm{poly}}({X_1}^L{X_2},{X_1}^L,\tau )$, where ${\rm{poly}}(a,b,c)$ is a third-order polynomial whose three roots are $a$, $b$, and $c$.

With the increase of quantization level of the state and action spaces, the number of feasible actions and states increases dramatically, which costs more time slots $\xi$ in each episode for
the RL-SAUD scheme to converge. As the side-effect to reduce the performance loss caused by quantization error, the increase of action-state pairs makes the RL-based algorithm cost
more searching and computational overhead in random exploration at the early stage of the learning process. For the DRL-SAUD scheme, the computational complexity of float-point
calculation in the deep neural networks, which is measured by float-point operations per second (FLOPs), is the main contributor to the overall complexity.
In the devised architecture, a network model including a single convolutional neural network (Conv) layers with ${C_{\rm{i}}}$ input channels and ${C_{\rm{o}}}$ output channels, and two fully connected (FC)
layers is considered. Then, the computational complexity of the proposed DRL-SAUD scheme is derived by the following theorem.
\begin{theorem}
  In the training process of the DRL-SAUD algorithm with $\tau $ episodes and $\xi$ time slots for each episode, the computational complexity is given by
  \begin{equation}\label{sec4:h}\small
    {{\cal T}_2} = {\cal O}\left( {\tau \xi {C_{\rm{i}}}w_{\rm{i}}^2{C_{\rm{o}}}{{\left( {({w_{\rm{i}}} - v)/s + 1} \right)}^2}} \right)
  \end{equation}
  where $v$ and $s$ denote the size and the stride of the convolutional kernel, respectively; ${w_{\rm{i}}}$ denotes the size of each input channel of the Conv layer.

\end{theorem}
\indent\indent \emph{Proof.} See the details in Appendix C.\QEDA

$\textbf{Remark 3}$. The concept of open service in O-RAN enables manufacturers to expand the functions of their network designs, and mobile operators can also support the coexistence of multiple
  vertical services. In practical deployment of multiple heterogeneous services, there is a significant increase in the amount of input parameters to the neural networks.
  Thus, it is difficult to analyze huge amount of information of the environment and the system by merely using a simple network of FC layers.
  Moreover, numerous parameters in the FC layers slow down the computation and it is more likely to cause overfitting. By utilizing Conv layers,
  one can reuse the parameters of the convolution kernel without consuming too much computational complexity overhead as shown in this theorem.
  This helps better extract the high-dimensional features of the complex system and realize a more efficient data-driven intelligent scheme for the agent.
\begin{figure}[htbp]
  \centering
  \subfigure[]{
  \label{fig:f1}
  \includegraphics[width=3.5 in]{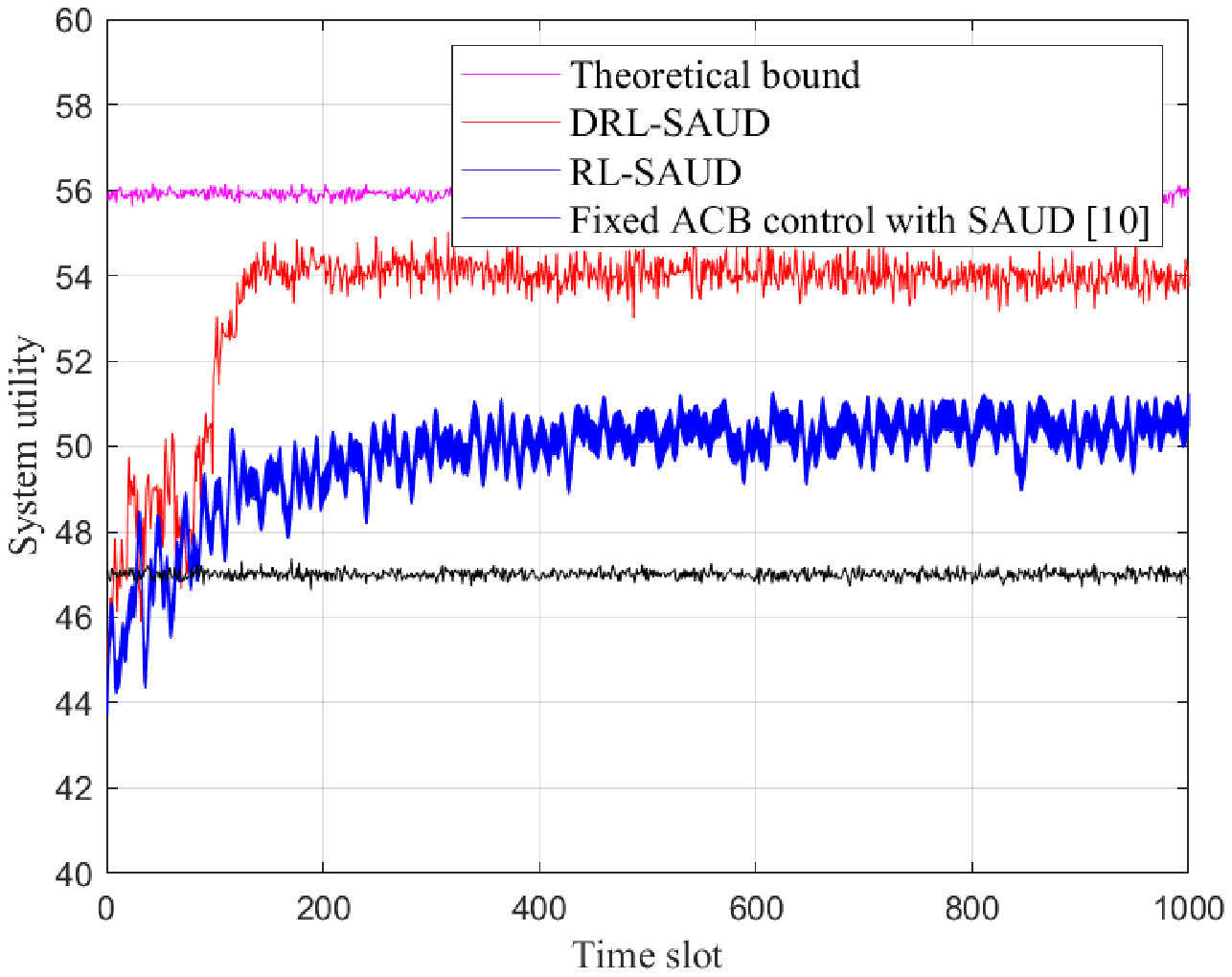}}
  \subfigure[]{
  \label{fig:f2}
  \includegraphics[width=3.5 in]{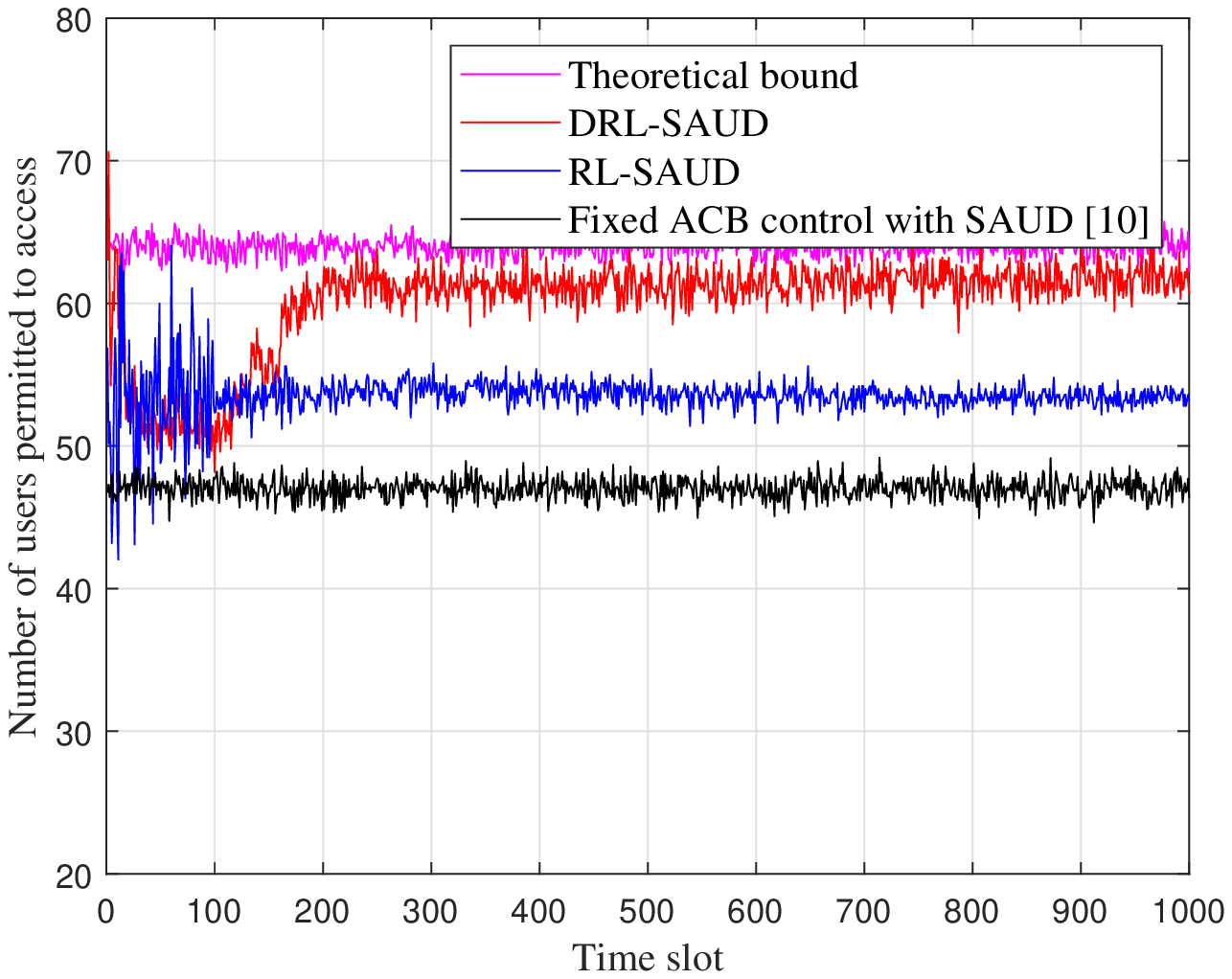}}
  \subfigure[]{
    \label{fig:f3}
    \includegraphics[width=3.5 in]{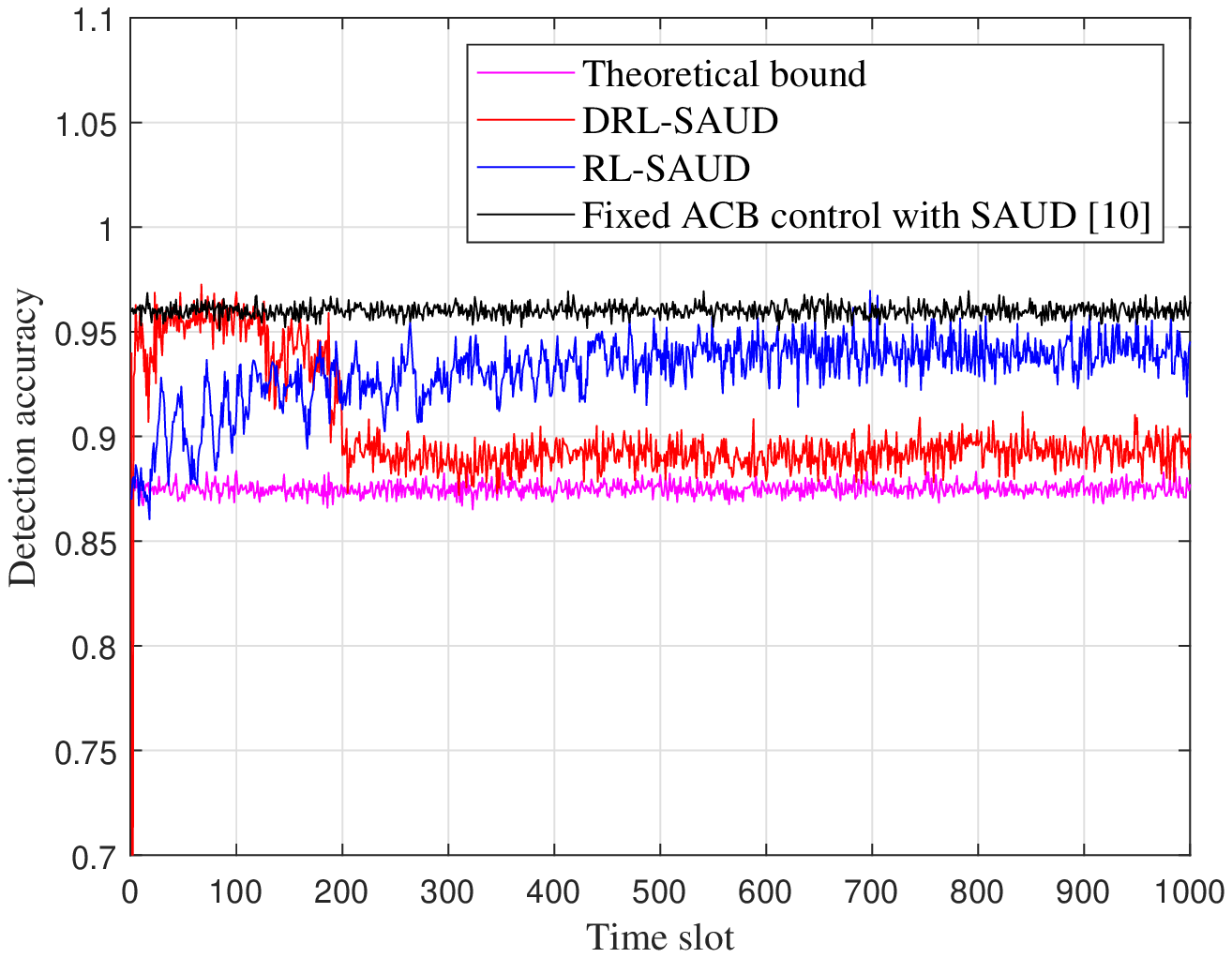}}
  \caption{Performance of the proposed RL-SAUD and DRL-SAUD schemes in
  (a) system utility,
  (b) access efficiency, i.e., number of users permitted to access, and
  (c) active user detection accuracy. An mMTC service with massive access requests from huge amount of users is considered and supported. A benchmark scheme, i.e. the fixed ACB factor with SAUD, is also depicted for comparison.}
  \label{figq}
\end{figure}
\begin{figure*}[htbp]
  \centering
  \subfigure[]{
  \label{fig:mu1}
  \includegraphics[width=3.5 in]{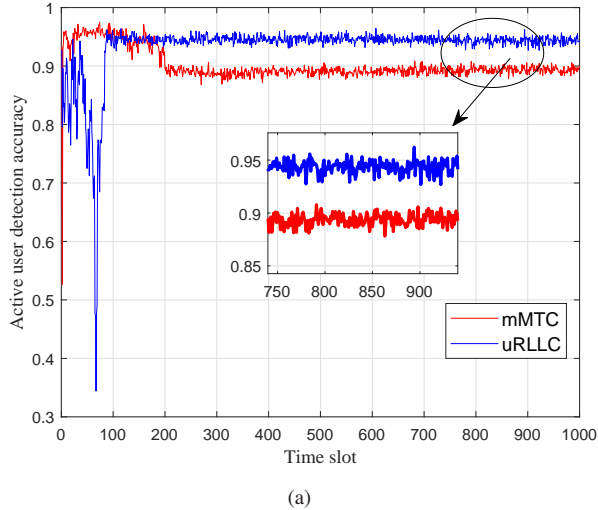}}
  \subfigure[]{
  \label{fig:mu2}
  \includegraphics[width=3.5 in]{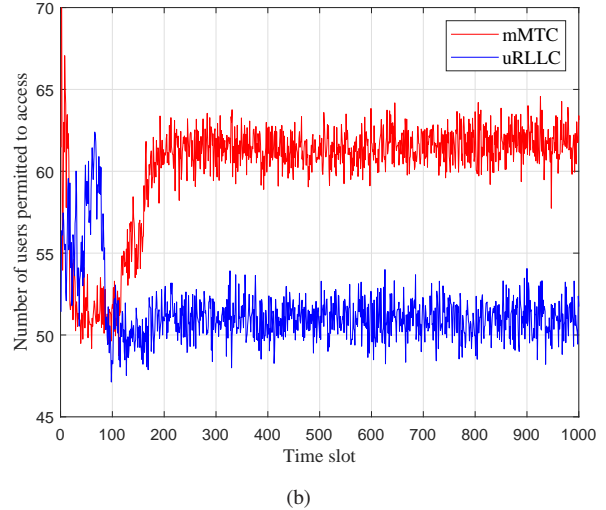}}
  \caption{Performance of the proposed DRL-SAUD scheme applied in a uRLLC service compared with that of an mMTC service:
  (a) Active user detection accuracy;
  (b) Number of users permitted to access the network.}
  \label{figq}
\end{figure*}

\vspace{0.1in}
\section{Simulation Results}
In this section, the performance of the proposed RL-SAUD and DRL-SAUD schemes is evaluated through extensive simulations. Some typical metrics,
such as the number of users permitted to access and the active user detection accuracy, are investigated to show the performance of the massive random access control schemes.
The number of users permitted to access the network is investigated to show the performance of access efficiency and throughput of the users,
while the active detection accuracy is investigated to show the reliability and stability of the access requests.
The effectiveness and adaptability of the proposed schemes are validated in different scenarios and various heterogeneous vertical services, such as mMTC and uRLLC services.
The tendence of the access control strategy determined by the proposed schemes is also demonstrated for different classes of users with different access priorities.

The simulation configuration is set up as follows: The number of potential users residing within the network is set to $N=256$. The number of antennas of the gNB is $M=128$.
The carrier central frequency is located at $2 GHz$, and the number of subcarriers is $K = 1024$. The parameter configuration of the RL-SAUD scheme is listed in {\bf{Table 1}}.
For the DRL-SAUD scheme, the learning rate of the Critic applied for TD3 is set as $0.00001$, and the learning rate of the Actor is set as $0.00005$ due to delayed update.
The Actor and the Critic share an identical network architecture, which consists of one Conv layer and two FC layers. The capacity of the replay buffer is set as $1000$.
The size of a mini-batch is $128$. The variance of the additive noise to encourage exploration is set as $\epsilon _1=0.15$. The variance of the policy noise for the Target Actor is set as $\epsilon _2=0.25$,
whose clip boundary $g$ is set as $0.4$. The delayed update time is $d=4$.


  \begin{table}[b]
    \renewcommand{\arraystretch}{1.2}
    \caption{Parameter configuration for RL-SAUD}
    \vspace{-0.25cm}  
    \setlength{\abovecaptionskip}{-0.20cm}   
    \setlength{\belowcaptionskip}{-0.1cm}   
    \centering
    \small
    \begin{tabular}{l c}
  \toprule
    \qquad $\textbf{Parameters}$&$\textbf{Value}$\\
    \hline
    Number of quantization levels for ACB factor $X_1$ & $5$ \\
  Number of classes with different access priority $L$ & $2$ \\
  Number of quantization levels for detection accuracy $X_2$ & $5$ \\
  Learning rate $\varpi$  & $0.98$\\
  Discount rate $\beta $ & $0.3$ \\
  Initial $\epsilon $  & $0.9$  \\
  Final $\epsilon $ & $0.1$ \\
  \bottomrule
    \end{tabular}\label{tableSimul}
    \end{table}

The performance of the proposed RL-SAUD and DRL-SAUD schemes in the system utility, access efficiency, and detection accuracy is reported in Fig. 4,
where an mMTC service with massive access requests from huge number of users is considered. A fixed ACB control scheme with SAUD \cite{t10} is evaluated as a benchmark,
and the theoretical bound is also depicted for comparison. First, the performance of system utility is reported in Fig. 4(a). It can be observed from Fig. 4(a) that,
the proposed RL-SAUD and DRL-SAUD schemes significantly outperform the benchmark scheme with fixed ACB control, which validates the effectiveness of the proposed RL-assisted
mechanism in achieving a better access control utility for massive access scenarios. It is also demonstrated that the DRL-SAUD scheme is approaching the theoretical bound of
system utility, which verifies the superior performance of the DRL-assisted architecture in case of complex environments and high-dimensional state and action spaces.
The performance gain of DRL-SAUD over RL-SAUD validates the effectiveness of utilizing the TD3 architecture with deep neural networks to extract more complex information from
the environment, and the degradation on the RL-based scheme caused by quantization error.

\begin{figure*}[htbp]
  \centering
  \subfigure[]{
  \label{fig:ff1}
  \includegraphics[width=3.5 in]{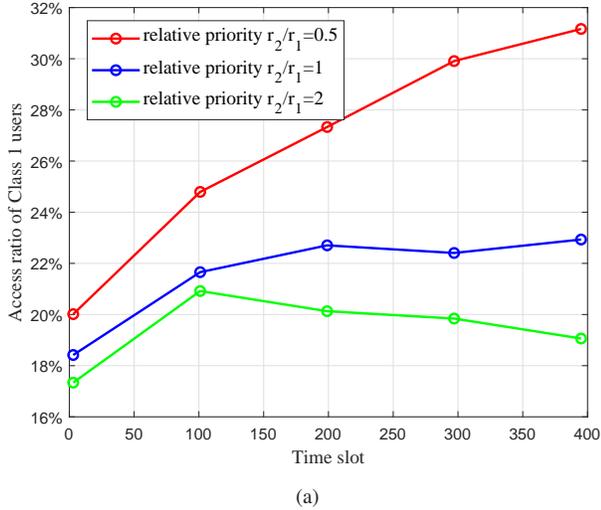}}
  \subfigure[]{
  \label{fig:ff2}
  \includegraphics[width=3.5 in]{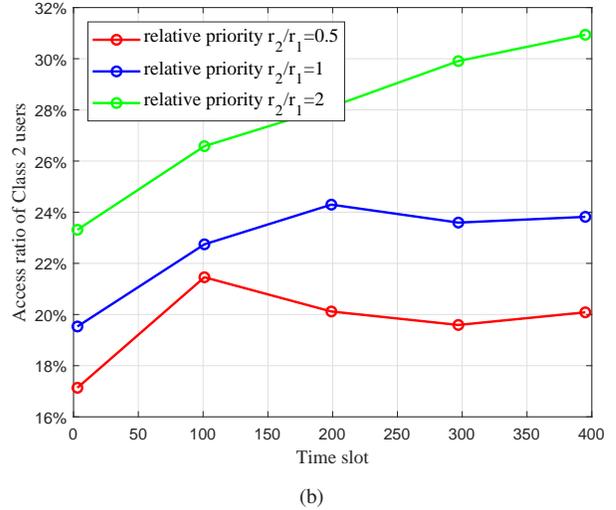}}
  \caption{The performance of access ratio, i.e. the percentage of users permitted to access the network, for two classes with different access priority scores:
  (a) Access ratio of Class 1 users with access priority score $r_1$;
  (b) Access ratio of Class 2 users with access priority score $r_2$.}
  \label{figq}
\end{figure*}

Second, the performance of access efficiency, which is indicated by the number of users permitted to access the network, is reported in Fig. 4(b).
It is observed from Fig. 4(b) that, DRL-SAUD permits about $62$ users to access the network, while RL-SAUD permits about $53$ users to access. Thus, DRL-SAUD can support more users than RL-SAUD
because it overcomes the bottleneck of quantization error and can find a better solution approaching the optimal bound, which is favorable for mMTC services with massive users
intended to access. In comparison, the benchmark scheme with fixed ACB only permits $48$ users to access, which shows its lack of flexibility to massive access requests.
Since the value of the fixed ACB factor is set empirically, it is difficult to reach a suitable setting in realistic complex and time-varying environments.

Third, the active user detection accuracy is reported in Fig. 4(c). It is shown from Fig. 4(c) that, the user detection accuracy of RL-SAUD and DRL-SAUD reaches approximately $94.93\%$ and $88.07\%$,
respectively. Although RL-SAUD has a relatively higher detection accuracy than DRL-SAUD, however, the number of permitted-access users of RL-SAUD is much fewer than that of DRL-SAUD,
which leads to a lower system utility as reported in Fig. 4(a). This implies that a bit decrease in detection accuracy can be compensated by a great increase in the number of
permitted-access users. Note that, the intelligent agent bears in its mind that maximizing the system utility function as given in (4) is its goal. Therefore,
an optimal trade-off in between should be pursued in order to obtain a higher system utility, and this best trade-off strategy is just the solution that the DRL-SAUD scheme
is searching for and finally converges to.

In addition, it can also be observed from Fig. 4 that, a period of action exploration is carried out in the early stage of the learning process of the proposed schemes.
This implies that, in the early stage of the learning process, the RL-based methods try some random selections of ACB factors to probe the environment efficiently and coarsely
in a large scale, which is helpful for finding a better solution at the cost of a bit exploration overhead.

Next, in order to demonstrate the capability of the proposed schemes in supporting and adaptively switching between different heterogeneous services,
the performance of the DRL-SAUD scheme in a uRLLC service is evaluated, which is compared with that of an mMTC service, as reported in Fig. 5. As described in Section IV-B,
for the uRLLC service, the third term of the system utility function in (4) is activated by setting a positive value of the coefficient $\rho _2$, which plays a role of penalty
on the user detection error and thus encourages better reliability of connection. Specifically, for the uRLLC service in the simulations in Fig. 5, the coefficient is set as $\rho _2=100$.

The performance of active user detection accuracy for the DRL-SAUD scheme applied in an mMTC service and a uRLLC service is reported in Fig. 5(a). It is shown by the results in
Fig. 5(a) that, the detection accuracy of the uRLLC service is about $5\%$ higher than that of the mMTC service. This improvement is beneficial for the agent, i.e., the gNB,
to make a prompt and effective response to a user who initiates an access request in an uRLLC service. It can also be noted from Fig. 5(a) that,
in the early stage of the learning process, the curve of detection accuracy for the uRLLC service has a deep valley, but is then pulled up rapidly.
This indicates that the DRL-SAUD scheme firstly performs initial random exploration to probe the environment, and then can rapidly adjust its strategy and converge to an optimized
solution because of the influence of the penalty of detection accuracy on the system utility.

On the other hand, the performance of access quantity, i.e., the number of users permitted to access, for the mMTC and uRLLC service is reported in Fig. 5(b).
It is shown that the proposed scheme permits about $62$ users to access for the mMTC service, while about $51$ users are permitted to access for the uRLLC service.
This result implies that the proposed scheme aims to permit more users to access the network for the mMTC service, while it determines to sacrifice a bit of access quantity
to improve the reliability and stability of the access connections for the uRLLC service. In the proposed RL-based framework, the agent can easily switch from an mMTC service
with a policy favorable for a larger access quantity, to a uRLLC service with a policy favorable for accurate user detection and reliable access connection, simply by setting a
positive value of $\rho _2$ to include the second penalty term on detection error in the utility function in (4). In practical implementation in the O-RAN paradigm,
rapid switching between different heterogeneous services can be realized by simply adjusting the penalty coefficients in the xAPPs.

To observe the behavior of the proposed DRL-SAUD scheme when faced with users with different access priorities, the access ratio of two differently prioritized classes of users,
i.e., Class $1$ and Class $2$, is reported in Fig. 6. The access ratio of a certain class of users is defined as the percentage of the users permitted to access the network with
respect to all the potential users in that class. Let $r_2/r_1$ represent the relative priority between the two classes considered, which is defined as the ratio of the priority
score of Class 2 $r_2$ with respect to the priority score of Class 1 $r_1$. In this case, the first penalty term on the system utility in (4), i.e.,
the penalty due to the variance of the access quantities between different prioritized classes of users, is activated by setting the corresponding coefficient as $\rho _1=120$.

Specifically, the performance of access ratio for the users in Class $1$ and Class $2$ are reported in Fig. 6(a) and Fig. 6(b), respectively.
Three cases with different values of relative priority are investigated for comparison, i.e., $r_2/r_1$ is set as $0.5$, $1$, and $2$. From Fig. 6 (b), it is observed that the access ratio
of Class $2$ is $30.9\%$ in case of $r_2/r_1=2$, which is $9.39\%$ and $35.27\%$ higher than the cases of $r_2/r_1=1$ and $r_2/r_1=0.5$, respectively,
which indicates that the proposed scheme has learned the tendence to allow more higher-prioritized users to access the network to improve the system utility.
It is can also be noted by comparing Fig. 6(a) and Fig. 6(b) that, the access ratio of the two classes is similar to each other in case of $r_2/r_1=1$,
which implies that the proposed scheme has learned to permit approximately the same amount of users to access with the same access priority.
The results in Fig. 6 have verified the adaptability of the proposed DRL-assisted scheme to different prioritized users or various heterogeneous services with different QoS requirements.

\vspace{0.1in}
\section{Conclusion}
Faced with the challenge of massive random access control in the context of O-RAN, this paper has proposed an RL-assisted framework of dynamic access control,
which can be deployed in the near-RT RIC at the gNB. In order to preserve the sparsity of the access requests to guarantee the accuracy of SAUD in case of ultra-dense traffic,
the proposed RL-SAUD scheme can dynamically adjust the ACB control strategy in a closed-loop access control process. A system utility function, which is in favor of increasing
the quantity of the users permitted to access the network, has been devised and utilized to train the RL model, and two penalty terms related with the variance of access ratio
and the detection accuracy are adopted to support a proper tradeoff and flexible switching between different heterogeneous vertical applications, such as mMTC and uRLLC services.

Furthermore, in order to overcome the quantization error of the RL-based scheme due to discretizing the actions and states using, the DRL-SAUD scheme has been designed based on the
Actor-Critic underlying TD3 framework. The information of the environment can be better extracted by the deep neural networks, and the policy and actions can be chosen from a
continuous space to obtain an improved solution approaching the optimal bound. Past experiences are exploited to accelerate the convergence of learning by using experience replay
buffer. The theoretical analysis and simulation results have validated the efficiency, adaptability, and reliability of the proposed schemes in dynamic and intelligent massive
access control for different QoS requirements, different vertical services and different prioritized classes of users. The technique is promising to be applied in the xAPPs in the
O-RAN paradigm to provide an efficient and effective RIC for the ever-crowded and ever-complex radio access environments and services.

\begin{appendices}
  \section{Proof of Theorem 1}
  \emph{Proof.} According to related research in literature \cite{t44}, it has been proved that the active user detection accuracy $c$ is lower-boudnded by
\begin{equation}\label{sec4:h}\small
  c \ge 1 - {a_1}\exp \left( { - {a_2}{\rm{min}}\left\{ {{{\left\| {\bm{\alpha }} \right\|}_1},{\rm{ log}}(N - {{\left\| {\bm{\alpha }} \right\|}_1})} \right\}} \right),
    \tag{A.1}
\end{equation}
subject to the following constraint,
\begin{equation}\label{align:w0cons}\small
  \begin{split}
    &{M \ge 2\left( {{{\left\| {\bm{\alpha }} \right\|}_1} + \frac{1}{{\varepsilon _N^2}}} \right)\log (N - {{\left\| {\bm{\alpha }} \right\|}_1})}\\
    &{{{\left| {{\lambda _n}} \right|}_{\min }} > \vartheta {\varepsilon _N}}
  \end{split}
  \tag{A.2}
\end{equation}
where $N$, $M$, and $||{\bm{\alpha }}|{|_1}$ denote the number of all the potential users in the network, the measurement vector size, and the sparsity level of the active user requests.
According to the CS theory \cite{t5}, the sparsity level that can be recovered accurately should be smaller than the number of measurement data $M$, so we have $M > {\left\| {\bm{\alpha }} \right\|_{1,\max }}$. Then, we can derive that,
\begin{equation}\label{sec4:h}\small
  \begin{split}
    2(||{\bm{\alpha }}|{|_1} + \frac{1}{{\varepsilon _N^2}})\log (N - M) &\le 2(||{\bm{\alpha }}|{|_1} + \frac{1}{{\varepsilon _N^2}})\log (N - ||{\bm{\alpha }}|{|_1})\\
  {\rm{                                   }} &\le 2(||{\bm{\alpha }}|{|_1} + \frac{1}{{\varepsilon _N^2}})\log N.
  \end{split}
  \tag{A.3}
\end{equation}
Substituting the constraint in (15) into (A.3), we have
\begin{equation}
  \begin{split}
  M &= 2\left( {{{\left\| {\bm{\alpha }} \right\|}_{1,\max }} + \frac{1}{{\varepsilon _N^2}}} \right)\log \left( {N - {{\left\| {\bm{\alpha }} \right\|}_{1,\max }}} \right)\\
  {\rm{    }} &\le 2\left( {{{\left\| {\bm{\alpha }} \right\|}_{1,\max }} + \frac{1}{{\varepsilon _N^2}}} \right)\log \left( N \right)\\
  {\rm{    }} &\le 2\left( {{{\left\| {\bm{\alpha }} \right\|}_{1,\max }} + \frac{M}{{2\log N}}\frac{1}{\varphi }} \right)\log \left( N \right),
\end{split}
  \tag{A.4}
\end{equation}
which is equivalent to
\begin{equation}\label{sec4:h}\small
  {\left\| {\bm{\alpha }} \right\|_{1,\max }} \ge \frac{{\eta N(1 - \frac{1}{\varphi })}}{{2\log N}}.
  \tag{A.5}
\end{equation}

Therefore, if the constraint in (14) is satisfied, the lower-bound of the detection accuracy can be derived in (A.1), which concludes the proof. \QEDA

\section{Proof of Theorem 2}
\emph{Proof.} If the access priority score of each class is identical, or equivalently, the number of user classes is only one, i.e., $L=1$, thus the total number of potential users in the network
and the ACB factor can be denoted by $N = {N_1}$ and $p = {p_1}$, respectively. Since there is no difference in ACB factors assigned to different classes,
the first penalty term of the utility function in (4) disappears, which is given by
\begin{equation}
  \begin{split}
    u &= c{p_1}{r_1}{N_1} - {\rho _2}(1 - c) = cprN - {\rho _2}(1 - c)\\
  &= f\left( p \right)prN - {\rho _2}\left( {1 - f\left( p \right)} \right)\\
  &= f\left( p \right)\left( {prN + {\rho _2}} \right) - {\rho _2},
\end{split}
  \tag{B.1}
\end{equation}
where the ACB factor $p \in [0,1]$, and the function $f\left( p \right)$ is defined by a monotonically decreasing convex curve \cite{t24} with properties given by
\begin{equation}
  \begin{split}
    &\frac{{\partial f\left( p \right)}}{{\partial p}} \le 0,\frac{{{\partial ^2}f\left( p \right)}}{{\partial {p^2}}} < 0,\\
    &{\left. {\frac{{\partial f\left( p \right)}}{{\partial p}}} \right|_{p \to {0^ + }}} \to {0^ - },\\
    &{\left. {f\left( p \right)} \right|_{p \to {1^ - }}} \to {0^ + }.
\end{split}
  \tag{B.2}
\end{equation}
Hence, maximizing the system utility function as given in (B.1) is a convex optimization problem.
It can be verified that the first derivative of the system utility $u$ with respect to the action $p$ satisfies
\begin{equation}
  \begin{split}
  {\left. {\frac{{\partial u}}{{\partial p}}} \right|_{p \to {0^ + }}} &= \frac{{\partial f\left( p \right)}}{{\partial p}}\left( {prN + {\rho _2}} \right) + rNf\left( p \right) = rNf({0^ + }) > 0,\\
  {\left. {\frac{{\partial u}}{{\partial p}}} \right|_{p \to {1^ - }}} &= \frac{{\partial f\left( p \right)}}{{\partial p}}\left( {prN + {\rho _2}} \right) + rNf\left( p \right) \\
  &= \frac{{\partial f\left( p \right)}}{{\partial p}}\left( {rN + {\rho _2}} \right) < 0.
\end{split}
  \tag{B.3}
\end{equation}
The second derivative of the system utility $u$ with respect to the action $p$ is strictly negative, which is as given by
\begin{equation}
  \frac{{{\partial ^2}u}}{{\partial {p^2}}} = \frac{{{\partial ^2}f\left( p \right)}}{{\partial {p^2}}}\left( {prN + {\rho _2}} \right) + \frac{{\partial f\left( p \right)}}{{\partial p}}(2rN + {\rho _2}) < 0.
  \tag{B.4}
\end{equation}
Therefore, it can be derived that the first derivative $\frac{{\partial u}}{{\partial p}}$ has a unique zero solution within the feasible range of $p \in [0,1]$.
Consequently, the convex function $u$ with respect to $p$ has a maximum value in the interval $p \in [0,1]$. \QEDA

\section{Proof of Theorem 3}
\emph{Proof.} For the convolutional neural network (Conv) layer in the architecture as shown in Fig. 3, the input data can be reshaped into a high-dimensional tensor of size ${C_{\rm{i}}} \times {w_{\rm{i}}} \times {w_{\rm{i}}}$,
where ${C_{\rm{i}}}$ is the number of input channels of the Conv layer. When the Conv layer has ${C_{\rm{o}}}$ output channels with each channel equipped with a convolution kernel of
size $v \times v$ and stride $s$, the number of float-point operations consumed by each Conv layer is calculated by
\begin{equation}
  {N_{{\rm{FLOPs}}}}({\rm{Conv}}) = {C_{\rm{i}}}w_{\rm{i}}^2{C_{\rm{o}}}w_{\rm{o}}^2
  \tag{C.1}
\end{equation}
where ${w_{\rm{o}}}$ is the size of each output channel of the Conv layer. According to related research in literature \cite{t46},
in the zero padding mode, the value of ${w_{\rm{o}}}$ is related with the input size, convolution kernel size, and stride as given by
\begin{equation}
  {w_{\rm{o}}} = \frac{{{w_{\rm{i}}} - v}}{s} + 1
  \tag{C.2}
\end{equation}
where the influence of bias is reflected by adding one to the right of (C.2). Since the FLOPs of the remaining two fully connected network (FC) layers is much smaller
than that of the Conv layer so that it can be neglected, so we have the computational complexity of each time slot of the DRL-based scheme as given by
\begin{equation}
  {N_{{\rm{FLOPs}}}}({\rm{time \ slot}}) =  {C_{\rm{i}}}w_{\rm{i}}^2{C_{\rm{o}}}{\left( {\frac{{{w_{\rm{i}}} - v}}{s} + 1} \right)^2}
  \tag{C.3}
\end{equation}
Finally, multiplying the number of FLOPs for each time slot of the DRL-based scheme with the number of episodes and time slots, we can derive (17). \QEDA
\end{appendices}

\bibliography{VLC_PU}
\bibliographystyle{IEEEtr}

\end{spacing}
\end{CJK}
\end{document}